\documentclass[10pt,twocolumn,letterpaper]{article}
\usepackage{multirow}
\usepackage{makecell}
\usepackage{amssymb}
\usepackage{amsmath}
\usepackage[pagenumbers]{cvpr} 

\usepackage{tcolorbox}
\usepackage{listings}
\usepackage{xcolor}

\definecolor{promptbg}{RGB}{250,250,250}
\definecolor{promptborder}{RGB}{180,180,180}

\lstdefinestyle{jsonstyle}{
  basicstyle=\ttfamily\footnotesize,
  backgroundcolor=\color{promptbg},
  frame=single,
  rulecolor=\color{promptborder},
  breaklines=true,
  showstringspaces=false,
  columns=flexible,
  keepspaces=true,
  stringstyle=\color{teal},
}
\definecolor{cvprblue}{rgb}{0.21,0.49,0.74}
\usepackage[pagebackref,breaklinks,colorlinks,allcolors=cvprblue]{hyperref}

\def\paperID{} 
\def\confName{CVPR}
\def\confYear{2026}

\title{Discover, Segment, and Select: A Progressive Mechanism for Zero-shot Camouflaged Object Segmentation}

\author{Yilong Yang, Jianxin Tian, Shengchuan Zhang, Liujuan Cao\thanks{Corresponding author}\\
Key Laboratory of Multimedia Trusted Perception and Efficient Computing, \\
Ministry of Education of China, Xiamen University, 361005, P.R. China.\\
{\tt\small \{Yilong.Yang, zsc\_2016, caoliujuan\}@xmu.edu.cn, thetian666@stu.xmu.edu.cn}
}

\begin{document}
\maketitle
\begin{abstract}
Current zero-shot Camouflaged Object Segmentation methods typically employ a two-stage pipeline (discover-then-segment): using MLLMs to obtain visual prompts, followed by SAM segmentation. However, relying solely on MLLMs for camouflaged object discovery often leads to inaccurate localization, false positives, and missed detections. To address these issues, we propose the \textbf{D}iscover-\textbf{S}egment-\textbf{S}elect (\textbf{DSS}) mechanism, a progressive framework designed to refine segmentation step by step. The proposed method contains a Feature-coherent Object Discovery (FOD) module that leverages visual features to generate diverse object proposals, a segmentation module that refines these proposals through SAM segmentation, and a Semantic-driven Mask Selection (SMS) module that employs MLLMs to evaluate and select the optimal segmentation mask from multiple candidates. Without requiring any training or supervision, DSS achieves state-of-the-art performance on multiple COS benchmarks, especially in multiple-instance scenes.
\end{abstract}

\section{Introduction}
The human visual system exhibits a remarkable ability to identify objects that seamlessly blend into their surroundings, a capability that has long inspired research in computer vision. Camouflage Object Segmentation (COS) aims to computationally replicate this ability, with critical applications in medical diagnosis~\cite{fan2020pranet}, agricultural monitoring~\cite{rustia2020application, wang2024depth}, autonomous driving~\cite{martinez2025mitigation} and military surveillance~\cite{haider2025identification, liu2025multi}. Despite its significance, COS is hampered by a reliance on large-scale annotated datasets and specialized models, which limit scalability and generalization to diverse real-world scenarios.

Recent advances in multimodal large language models (MLLMs)~\cite{bai2025qwen2,lu2024deepseek,achiam2023gpt,chen2023shikra,liu2023visual} and vision foundation models like SAM~\cite{kirillov2023segment,ravi2024sam,tang2024chain} have shown that strong visual reasoning, localization and segmentation can be achieved without any task-specific training. By combining language-guided semantic priors from MLLMs with promptable SAM, it becomes feasible to perform COS in a training-free, zero-shot manner. In general, the pipeline mainly involves two steps. The first step is using an MLLM, or coupled with a Vision Language Model (VLM) such as CLIP~\cite{radford2021learning}, to generate location prompts (\eg, bounding boxes, points) that locate the potential camouflaged objects (discovery stage). The second step is feeding these prompts into SAM to produce segmentation masks (segmentation stage). Following this paradigm, several recent works~\cite{hu2024relax,hu2024leveraging,tang2024chain,yin2025stepwise,yin2025simple} have demonstrated promising zero-shot COS performance.
\begin{figure}[!t]
    \centering
    \includegraphics[width=\linewidth]{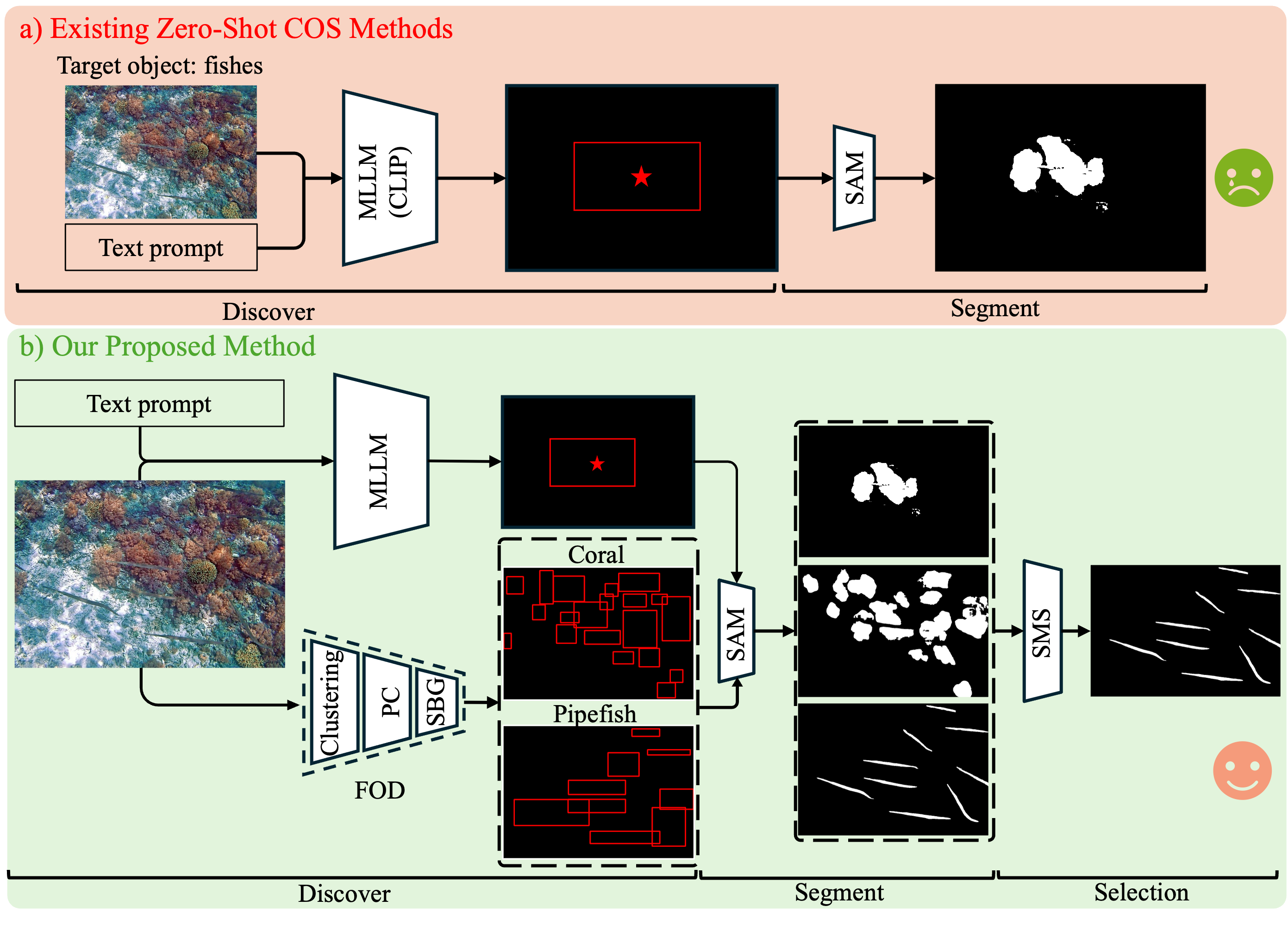}
    \caption{Comparison between the proposed DSS framework and prior zero-shot COS methods.}
    \label{fig:our_vs_prev}
\end{figure}
However, existing MLLM–SAM pipelines still struggle in camouflage scenarios: (1) MLLMs often yield inaccurate or missed localizations due to a reliance on high-level semantics over fine-grained cues, which misleads SAM and causes segmentation errors. (2) Their performance degrades notably in multi-instance scenes, where spatially distinct camouflage objects exist. These challenges highlight a critical limitation: relying solely on MLLMs is insufficient for generating high quality visual prompts.

To overcome these limitations, we propose a novel Discover, Segment, and Select pipeline (DSS)—that not only fundamentally redefines the discovery process but also introduces a reasoning-based selection stage for zero-shot COS. As shown in~\cref{fig:our_vs_prev}, Previous approaches either require MLLMs to perform localization beyond their intended capability or rely on CLIP-based prompt refinement, which is suboptimal for dense prediction. In contrast, our DSS framework introduces a Feature-coherent Object Discovery (FOD) module and a reasoning-based selection stage, enhancing robustness and segmentation quality in zero-shot COS. In the Discover stage, we leverage both semantic cues from MLLMs and structural priors from visual feature clustering to generate diverse and high-quality region proposals. These proposals are then passed to SAM in the Segment stage, producing multiple fine-grained candidate masks. Finally, in the Select stage, the MLLM serves as a reasoning-based selector, evaluating all candidates in the visual context to identify the most semantically and structurally consistent segmentation result. This tri-stage design enhances robustness against incomplete, inaccurate and incorrect proposals, therefore improves mask quality under a fully zero-shot, training free setting.

Experiments demonstrate that the proposed DSS framework achieves outstanding performance in zero-shot COS. The key contributions are summarized as follows:
\begin{itemize}
  \item We propose a Discover, Segment, and Select pipeline to address the prevalent issues of inaccurate localization in zero-shot COS. This is achieved by augmenting the discovery process with visual clustering and introducing a final selection stage for optimal mask identification.
  \item We introduce a Part Composition (PC) module that integrates discrete object parts, effectively enhancing the coherence and completeness of the segmentation output for complex camouflaged objects.
  \item We propose Similarity-based Box Generation (SBG), a robust bounding box generation method designed for multiple-instance scenarios. Its core strength lies in preventing the omission of instances, which ensures the preservation of all instances and directly leads to superior segmentation accuracy.
  \item We design a Semantic-driven Mask Selection (SMS) module that employs MLLMs for selection among candidate masks, ensuring optimal final segmentation.

\end{itemize}
\section{Related Work}
\label{sec:Related_work}
\subsection{Camouflaged Object Detection}
Camouflaged Object Segmentation~\cite{fan2020camouflaged,fan2021concealed} aims to identify and segment objects that seamlessly blend into their surroundings, presenting significant challenges due to their low foreground-background contrast.
With the advent of deep learning, convolutional neural networks have become the dominant paradigm for COS. Notable methods include SINet~\cite{fan2020camouflaged}, which introduced a search identification mechanism to enhance feature representation, and more recent works like ZoomNet~\cite{pang2024zoomnext} and CGCOD~\cite{zhang2024cgcod}, which leverage multi-scale features and context-aware modules to improve segmentation accuracy. However, these methods are fully supervised and require large annotated datasets, limiting their scalability and generalization to diverse real-world scenarios. To overcome the limitations of annotation dependency, the field has progressively shifted towards more flexible paradigms. Initial efforts in unsupervised learning explored techniques like cross-domain adaptation~\cite{zhang2023unsupervised} to learn patterns without COS-specific labels. However, it still require a training process on (often synthetic or related) data, and their performance is constrained by the domain gap and the quality of the auxiliary data. 
More recently, zero-shot approaches leverage pre-trained models and multimodal large language models (MLLMs) reasoning to perform COS without any task-specific training~\cite{hu2024relax,hu2024leveraging,yin2025stepwise,hao2025simple,du2025upgen,lei2025towards}. While these zero-shot methods offer greater flexibility and generalization, they often struggle with the inherent ambiguity of camouflaged scenes, leading to challenges in accurate localization and segmentation.
\begin{figure*}[t]
    \centering
    \includegraphics[width=\linewidth]{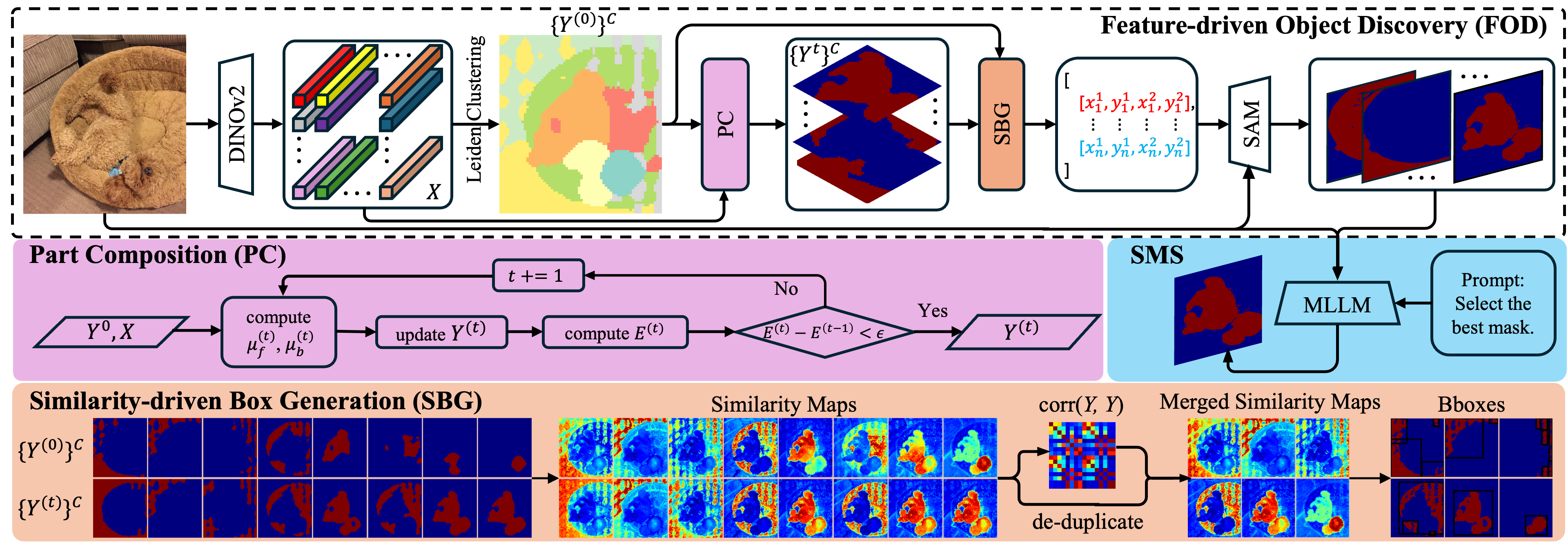}
    \caption{The overall pipeline of the Discovery-Segment-Selection Framework. The framework operates in three stages: (a) Feature-guided Object Discovery (FOD): the Part Composition (PC) module refines low-resolution clustering masks, followed by a Similarity based Box Generation (SBG) module to produce prompt boxes; (b) The Segment Anything Model generates segmentation mask based on the input prompts. (c) Semantic-driven Mask Selection (SMS): an MLLM evaluates all candidate masks and selects the final segmentation. 
    }
    \label{fig:overall_pipeline}
\end{figure*}
\subsection{Prompt-based Zero-Shot Segmentation}
\label{sec:intro}
The recent emergence of powerful foundation models, particularly the Segment Anything Model (SAM)~\cite{kirillov2023segment,ravi2024sam}, has catalyzed a paradigm shift in segmentation towards prompt-driven, zero-shot inference. A predominant strategy involves harnessing the rich prior knowledge within MLLMs to generate semantic-aware prompts such as points, bboxes or coarse masks for SAM. As an example, GenSAM \cite{hu2024relax} commences by utilizing BLIP2~\cite{li2023blip} to generate the class label of camouflaged objects. Coupled with this, CLIP Surgery~\cite{li2023clip} is employed to derive class attention maps, which in turn guide the generation of point prompts. These points are then fed into SAM to obtain the final segmentation masks. Following this paradigm, ProMaC~\cite{hu2024leveraging} improves GenSAM by exploiting hallucination priors to reduce the inaccuracy of MLLM-generated text prompts. MMCPF~\cite{tang2024chain} enhances initial coarse bboxes with DINOv2~\cite{oquab2023dinov2} features to refine points prompts for SAM. RDVP-MSD~\cite{yin2025stepwise} introduces a stepwise decomposition strategy to resolve semantic ambiguity in captions, coupled with a region-constrained approach that samples point prompts from coarse MLLM localizations to help SAM focus on local textures and boost accuracy. IAPF~\cite{yin2025simple} leveraging Grounding DINO ~\cite{liu2024grounding} to produce instance-level bbox prompts, alongside a single-foreground multi-background prompting strategy to sample region-constrained point prompts within each box, thus enhancing SAM's ability to handle multi-instance scenarios. 

Nonetheless, prompt-based pipelines inherit several failure modes from their constituent models. When MLLMs produce inaccurate or incorrect localizations, the SAM is prone to under-segmentation, over-segmentation, or entirely missing objects. Moreover, in multi-instance scenes, performance often degrades because language or coarse visual priors may focus on dominant instances and neglect others. A common thread in existing zero-shot camouflage segmentation approaches is their reliance on visual grounding cues from MLLMs and VLMs. This focus, however, largely neglects the rich, intrinsic visual discriminatory features present in the image, which are crucial for compensating for the lack of clear semantics and lead to suboptimal performance in visually deceptive scenarios.
\section{Method}
\subsection{Framework Overview}
The proposed framework consists of three core stages: Feature-coherent Object Discovery (FOD), promptable SAM segmentation, and Semantic-driven Mask Selection (SMS), as illustrated in \cref{fig:overall_pipeline}. Given an input image, the FOD module extracts patch-level features via a self-supervised encoder and groups them through unsupervised clustering. The Part Composition (PC) module further enhances cluster consistency, producing a set of low-resolution binary masks. A Similarity-based Box Generation (SBG) module then computes semantic affinity maps between foreground regions and all image patches, from which high-quality bbox prompts are extracted to guide the SAM in producing candidate masks. Finally, the SMS module evaluates all candidate masks using via MLLM and selecting the most appropriate one as the final segmentation.
\subsection{Feature-coherent Object Discovery}
The Discover stage begins with our Feature-coherent Object Discovery (FOD) module. To achieve more comprehensive object discovery, FOD leverages the self-supervised visual representation extracted from a pretrained vision encoder. Specifically, we obtain patch-level embeddings $\mathbf{X}=\{x_i\in \mathbb{R}^d\}^N$, where $ N=H^\prime\times W^\prime$ is the number of patches and $d$ is the feature dimension. These features capture rich semantic information, enabling effective differentiation between foreground and background regions. Then a clustering function $\mathcal{C}(\cdot)$ groups the patch-level feature map into $C$ clusters, yielding coarse binary masks $\{\mathbf{Y}_c\}^C$. To achieve semantically consistent and compact patch-level foreground-background separation, we iteratively refine these masks via feature coherence, enforcing intra-cluster similarity and inter-cluster separability. Each refined low-resolution mask is then converted to bboxes and used as prompt for segmentation. We break this process down into two steps: (1) Part Composition (PC) and (2) Similarity-driven Box Generation (SBG).\\
\noindent\textbf{Part Composition.} While clustering provides an initial grouping of features, it may over-segment a single camouflaged object into multiple parts. To address this, the PC module is introduced to refine the clustering results by enforcing semantic consistency among patches. Specifically, PC starts from the clustering results and iteratively refines the soft foreground probability of each patch by minimising a feature-coherence energy that encourages intra-class compactness and inter-class separability in feature space. Formally, we denote $\mathbf{Y}_c\in {0,1}^N$ as the initial foreground probability map of the $c$-th cluster. For notational simplicity, we drop the cluster index $c$ hereafter. At iteration $t\in\{1,\cdots,T\}$, $\mathbf{Y}^{(t)}=\{y_i^{(t)} \in [0,1]\}^N$ is updated as the relative distance to two centers and converted into soft values:
\begin{figure}[!t]
  \centering
   \includegraphics[width=\linewidth]{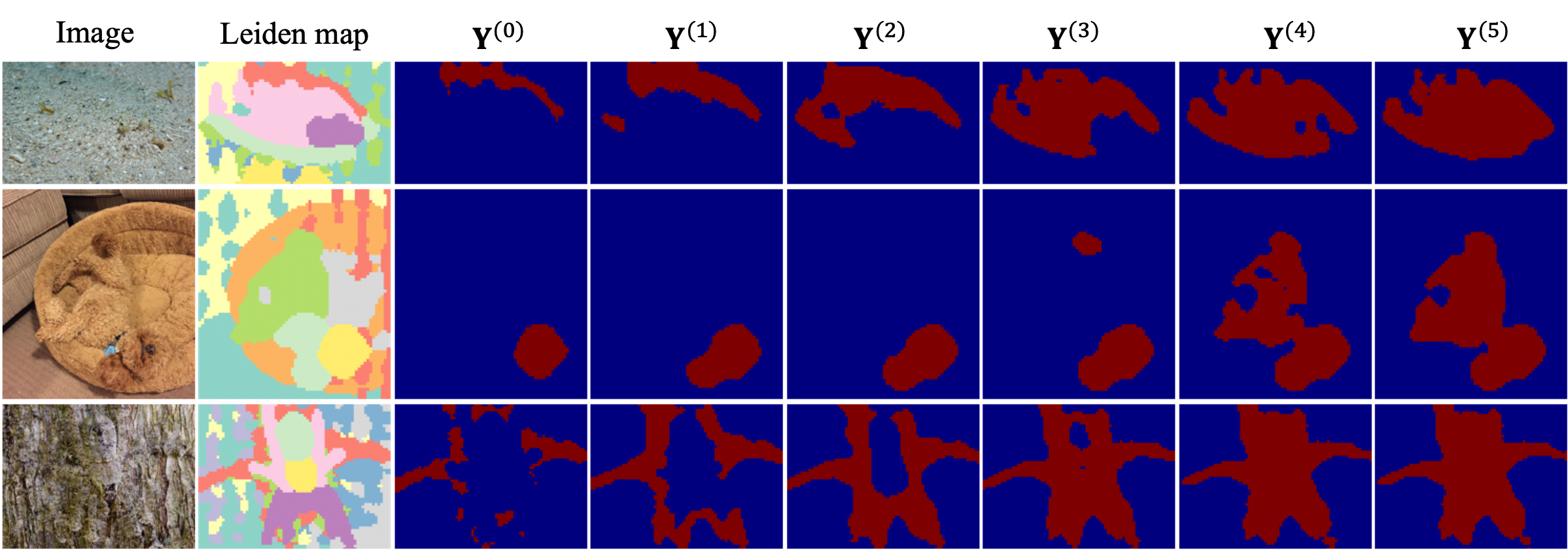}
   \caption{Visual tracking of the iterative refinement process. Left to right: the original image, Leiden clustering map, and binary maps from initial clustering $\mathbf{Y}^{(0)}$ to the final refined map $\mathbf{Y}^{(5)}$. Here we select one cluster from the clustering results for demonstration. The number of iterations may vary for different clusters until convergence.}
   \label{fig:iter_maps}
\end{figure}
\begin{equation}
y_i^{(t)} = \sigma( \|x_i - \mu_b^{(t-1)}\|_2 - \|x_i - \mu_f^{(t-1)}\|_2),
\label{eq:irfc_sigmoid}
\end{equation}
where $\sigma(\cdot)$ is the sigmoid function, $\mu_f^{(t)}$ and $\mu_b^{(t)}$ denote the foreground and background feature centroids at $t$-th iteration, respectively:
\begin{equation}
\mu_f^{(t)} = 
\frac{\sum_i y_i^{(t)} x_i}{\sum_i y_i^{(t)}}, 
\qquad
\mu_b^{(t)} = 
\frac{\sum_i (1 - y_i^{(t)}) x_i}{\sum_i (1 - y_i^{(t)})}.
\label{eq:irfc_centers}
\end{equation}
\cref{eq:irfc_sigmoid} updates each patch's foreground probability based on its relative distance to the foreground and background centroids, effectively pulling similar patches closer while pushing dissimilar ones apart. At the end of each iteration, the feature-coherence energy is evaluated:
\begin{equation}
\begin{aligned}
E^{(t)}& =\underbrace{
\Big(
\frac{1}{N_f^{(t)}}\sum_{i \in \mathcal{I}_f^{(t)}}\|x_i - \mu_f^{(t)}\|_2 + 
\frac{1}{N_b^{(t)}}\sum_{i \in \mathcal{I}_b^{(t)}}\|x_i - \mu_b^{(t)}\|_2
\Big)
}_{\text{intra-class compactness}} \\
&-\underbrace{\|\mu_f^{(t)} - \mu_b^{(t)}\|_2}_{\text{inter-class separation}},
\label{eq:irfc_energy}
\end{aligned}
\end{equation}
where $\mathcal{I}_f^{(t)}=\{i:y_i^{(t)}>0.5\}$ and $\mathcal{I}_b^{(t)}=\{i:y_i^{(t)}\leq0.5\}$ are the sets of foreground and background patches, with $N_f^{(t)}=|\mathcal{I}_f^{(t)}|$ and $N_b^{(t)}=|\mathcal{I}_b^{(t)}|$ being their respective cardinalities. The refinement stops automatically when the energy change falls below a threshold:
\begin{equation}
|E^{(t)} - E^{(t-1)}| < \epsilon,
\end{equation}
indicating convergence to a stable and coherent labeling configuration. The refinement is proceeded for each cluster and therefore resulting in $C$ masks $\{\mathbf{Y}_c^{(t)}\}^C$ which can be used to generate bbox prompts for the SAM. The flowchart of PC is illustrated in~\cref{fig:overall_pipeline}. An visual tracking of the iterative refinement process is shown in \cref{fig:iter_maps}. As observed, by progressively absorb similar neighboring patches into the same cluster while pushing dissimilar ones away, PC effectively refines the initial over-segmented masks into more coherent object-level segmentations.\\
\noindent\textbf{Similarity-driven Box Generation.} To generate bbox prompts for SAM from candidate masks ${\mathbf{Y}_c}^C$, we face two main challenges: (1) Directly extracting boxes from connected components may yield incomplete proposals, causing missed detection; and (2) Highly overlapping masks from the PC may produce redundant boxes for the same object. To overcome these issues, we first introduce a more robust prompt generation method based on self-similarity maps. For each mask $\mathbf{Y}_c$, we compute a self-similarity map that quantifies the semantic affinity in the feature space between its foreground region and all image patches:
\begin{equation}
    \label{equ:sim_map}
    \mathbf{sim}_c=sim(\mathbf{X},\mathbf{Y}_c)=\frac{\mu_f\cdot\mathbf{X}}{\|\mu_f\|\|\mathbf{X}\|}, \mu_f=\frac{\sum_i y_i x_i}{\sum_i y_i}.
\end{equation}
\begin{figure}[!t]
  \centering
   \includegraphics[width=\linewidth]{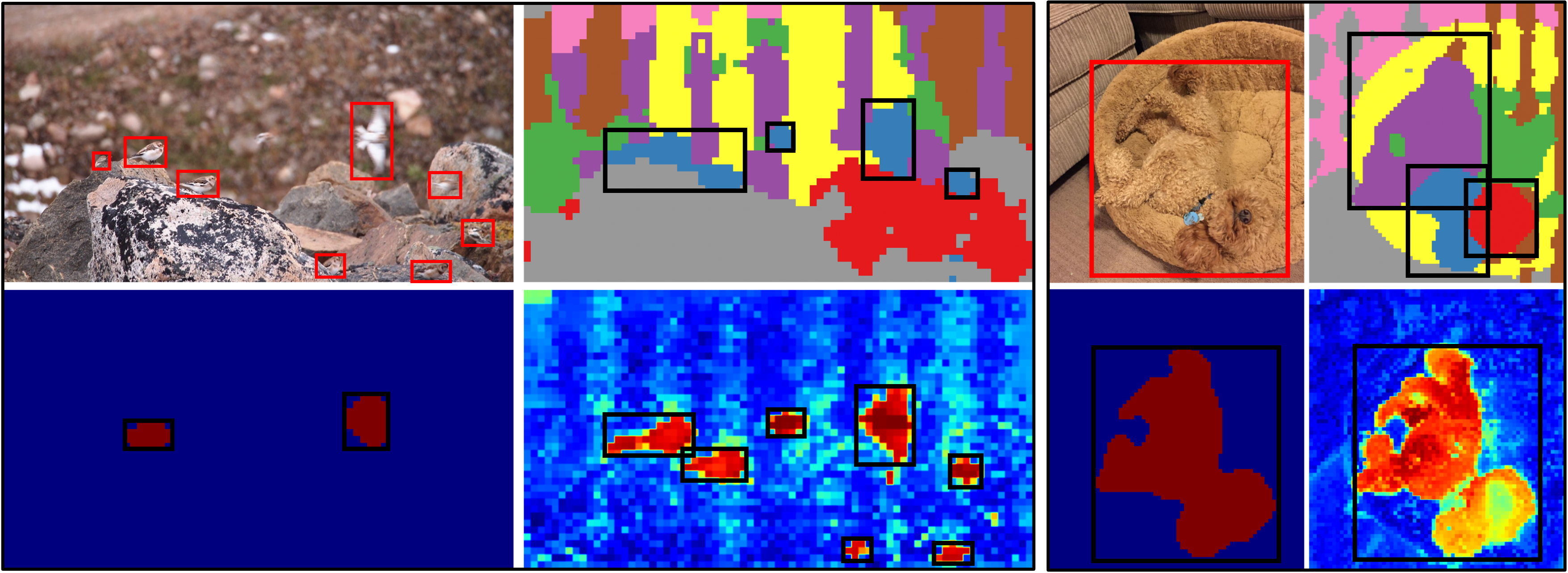}
   \caption{Visual comparison between bboxes generated from different strategies. In each 2$\times$2 grid, \textit{Top left}: bboxes from the ground truth mask. \textit{Top right}: bboxes from leiden clustered mask. \textit{Bottom left}: bboxes from PC refined mask. \textit{Bottom right}: bboxes from self-similarity map.}
   \label{fig:sim_vs_binary}
\end{figure}

To assess and mitigate redundancy among the similarity maps $\{\mathbf{Y}_1,\cdots,\mathbf{Y}_C\}$, we compute the pairwise Pearson correlation coefficients betwen all map pairs, resulting in a correlation matrix $\mathbf{R}\in\mathbb{R}^{C \times C}$, with each element $r_{i,j}$ quantifies the degree of linear dependency between maps $\mathbf{sim}_i$ and $\mathbf{sim}_j$.
Then a correlation threshold $\tau$ is subsequently applied to identify highly correlated pairs $(r_{i,j}>\tau)$, which are merged via averaging. This process effectively reduces the number of bbox proposals fed into the SAM, thus improving computation efficiency. Next, we threshold the self-similarity map to create a binary affinity map and extract bounding bboxes as prompts for segmentation. Then the final set of masks generated by SAM is denoted as:
\begin{equation}
\mathbf{M}_{\mathrm{FOD}} = \{ \mathrm{m}_1, \mathrm{m}_2, \dots, \mathrm{m}_{C^\prime}\}, \quad C^\prime\leq C.
\end{equation} 

The procedure is shown in the bottom of \cref{fig:overall_pipeline}. ~\cref{fig:sim_vs_binary} presents a visual comparison of bboxes generated from different strategies. As observed in~\cref{fig:sim_vs_binary}, bboxes derived from similarity maps exhibit higher completeness than binary masks, alleviating the risk of missing instances.
\subsection{Semantic-driven Mask Selection}
The final step of the DSS is to select the most plausible camouflaged object mask from the set of candidate masks $\mathbf{M}_\mathrm{FOD}$. Feeding all candidate masks into the MLLM simultaneously may gives the model excessive flexibility, which often triggers hallucination and results in incorrect selections. To alleviate this, we adopt a progressive pairwise comparison strategy guided by a heuristic scoring function. Specifically, we first assign each mask a confidence score $s_i$ based on its spatial consistency with the similarity map and its boundary contactness. The scoring function is defined as:
\begin{equation}
  \label{eq:score}
  s_i = \text{corr}(\text{m}_i, \textbf{sim}_i)+(1-\text{BC}(\text{m}_i)), \text{m}_i \in \mathbf{M}_{\mathrm{FOD}},
\end{equation}
where $\text{corr}(\text{m}_i, \textbf{sim}_i)$ denotes the Pearson correlation between mask $\mathrm{m}_i$ and the similarity map $\mathbf{sim}_i$. A higher correlation  indicates that the mask aligns well with the semantic similarity distribution of the image, implying stronger consistency with the potential camouflaged region. $\text{BC}(\mathrm{m}_i)$ represents the boundary contact rate, quantifying the proportion of mask edges touching the image boundary—since camouflaged objects rarely coincide with image borders, a higher $\text{BC}(\mathrm{m}_i)$ suggests that the mask likely includes background regions. The implementation of BC is detailed in the supplementary material~\cref{sec:impl_details_bc}.

We then rank these masks by their scores and retain the top-$K$ candidates. This step aims to prune implausible candidates in order to reduce subsequent pairwise comparisons, thus enhancing computational efficiency. The set $\{\text{m}_1, \ldots, \text{m}_K\}$ is then used for iterative pairwise selection. Starting from the two lowest-scoring masks, we iteratively feed the corresponding masked image regions (obtained via pixel-wise multiplication with the original image $I$) together with $I$ itself into the MLLM, accompanied by a query prompt $P_\mathrm{MMS}$ such as “Identify which masked image best corresponds to the camouflaged object in the original image.” The model outputs a categorical decision indicating the preferred mask:
\begin{equation}
\hat{m} = \text{MLLM}(I, {I \odot \mathrm{m}_a, I \odot \mathrm{m}_b}, P_\mathrm{SMS}).
\end{equation}
where $\odot$ denotes element-wise multiplication, and $\mathrm{m}_a$, $\mathrm{m}_b$ are the two masks under comparison. The winning mask $\hat{\text{m}}$ is then compared with the next higher-scoring mask in the sequence, repeating this for $K$-1 iterations until the final mask is determined. We empirically found that starting the pairwise comparison from low-score masks yields more stable outcomes. The effectiveness of the heuristic rules and the pairwise selection strategy, compared with the single-pass selection, is further validated in the supplementary material~\cref{sec:extend_ablation}.
\begin{table}[t]
  \caption{Hyperparameter setting.}
  \centering
  \footnotesize
  \begin{tabular}{ccl}
    \toprule
    HP  & Value & Remark\\
    \midrule
    $r$ & 0.5 & Leiden clustering resolution.\\
    $\epsilon$ & 1.0 & The energy change threshold.\\
    $\tau$ & 0.95 & Correlation threshold for de-duplication.\\
    $K$ & 5 & Top-K masks for MLLM mask selection.\\
    \bottomrule
  \end{tabular}
  \label{tab:hyperparameter}
\end{table}
\begin{figure*}[!t]
    \centering
    \includegraphics[width=\linewidth]{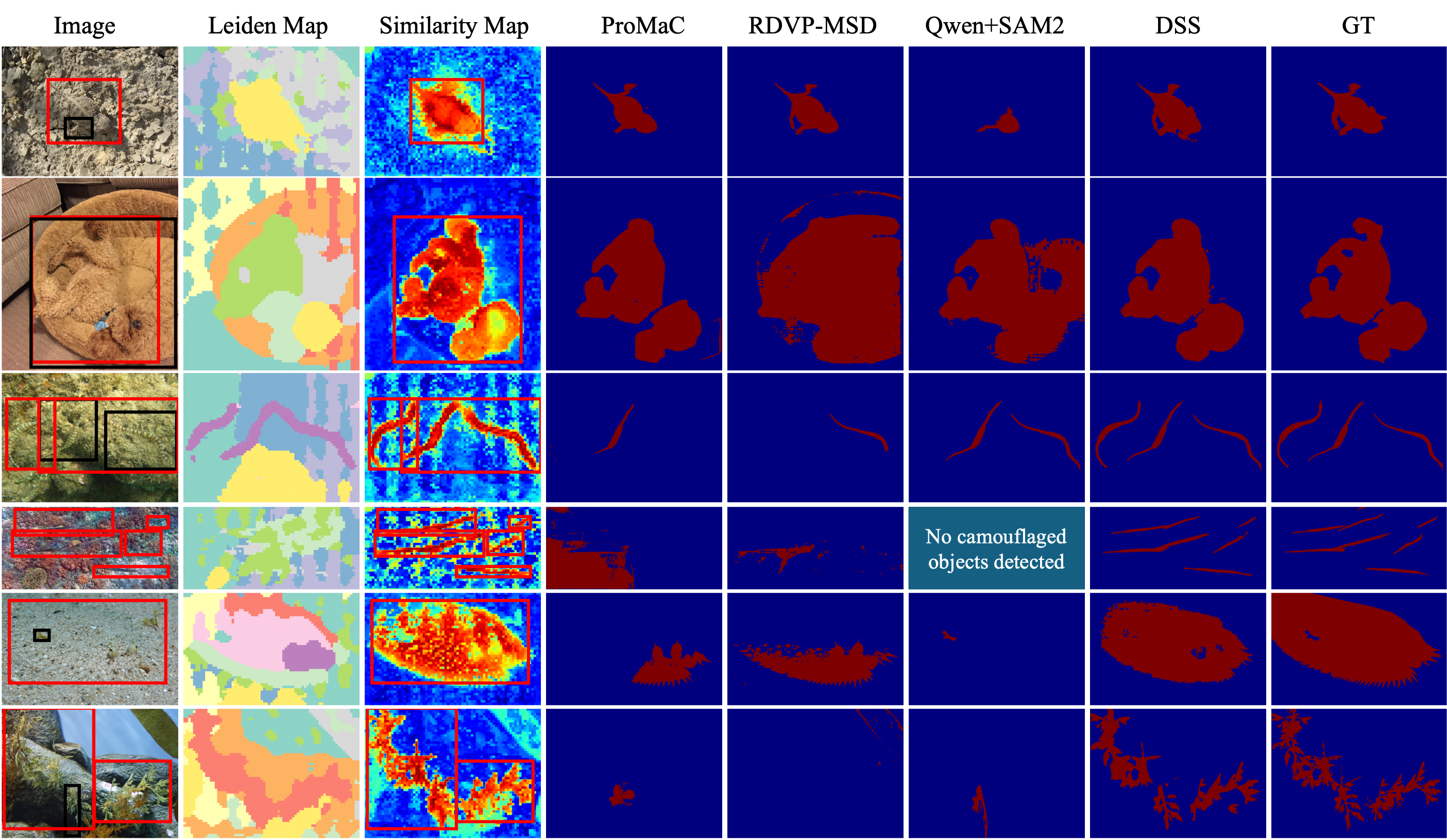}
    \caption{We present a visual comparison of our method against existing alternatives under challenging conditions. The red and black bounding boxes correspond to prompts generated by our approach and QWen, respectively. To aid interpretation, the third column visualizes the similarity maps from which our bounding boxes are derived.}
    \label{fig:pred_vis}
\end{figure*}
\section{Experiments}
\subsection{Experiment Settings}
\begin{table*}[ht]
\centering
\footnotesize
\caption{Quantitative comparison with state-of-the-art methods on four COD datasets “$\uparrow/\downarrow$” : the higher/lower the better. “-”: not available. “F”, “U” and “ZS” denote “Fully-supervised”, “Un-supervised” and “Zero-shot” method, repsectively. The best three results in zero-shot setting are highlighted in \textcolor{red}{red}, \textcolor{blue}{blue}, and \textcolor{green}{green}, respectively. The best results in “F”, “U” settings are in bold.}
\setlength{\tabcolsep}{2.5pt}
\begin{tabular}{l|c|c|cccc|cccc|cccc|cccc}
\toprule
\multirow{2}{*}{Methods} & \multirow{2}{*}{Venue} & \multirow{2}{*}{Sup} & \multicolumn{4}{c|}{CAMO(250)} & \multicolumn{4}{c|}{NC4K(4121)} & \multicolumn{4}{c|}{COD10K(2016)} & \multicolumn{4}{c}{CHAMELEON(87)} \\
 & & & $F_\beta^{w}\uparrow$ & $\mathcal{M}\downarrow$ & $S_\alpha\uparrow$ & $E_\phi\uparrow$
 & $F_\beta^{w}\uparrow$ & $\mathcal{M}\downarrow$ & $S_\alpha\uparrow$ & $E_\phi\uparrow$
 & $F_\beta^{w}\uparrow$ & $\mathcal{M}\downarrow$ & $S_\alpha\uparrow$ & $E_\phi\uparrow$
 & $F_\beta^{w}\uparrow$ & $\mathcal{M}\downarrow$ & $S_\alpha\uparrow$ & $E_\phi\uparrow$ \\
\midrule
HitNet \cite{hu2023high}& AAAI23 & F & .809 & .055 & .849 & .906 & .834 & .037 & .875 & .926 & .806 & .023 & .871 & .935 & .897 & .019 & .921 & .967\\
SENet \cite{hao2025simple} & TIP24 & F & .847 & .039 & .888 & .932 & .843 & .032 & .889 & .933 & .780 & .024 & .865 & .925 & .878 & .019 & .918 & .957\\
ZoomNeXt \cite{pang2024zoomnext} & TPAMI24 & F & .857 & .041 & .889 & .945 & .863 & .028 & .903 & \bf .951 & .827 & \bf .018 & \bf .898 & \bf .956 & .885 & .018 & .924 & \bf .975 \\
CGCOD \cite{zhang2024cgcod} & MM25 & F & .864 & \bf .036 & .896 & \bf .947 & .869 & \bf .026 & .904 & .949 & .824 & \bf .018 & .890 & .948 & \bf .902 & \bf .017 & \bf .931 & .972 \\
CAMF \cite{lei2025towards}  & TPAMI25 & F & \bf .887 & .041 & \bf .901 & .933 & \bf .897 & .033 & .905 & .933 & \bf .845 & .025 & .890 & .923 & .895 & .028 & .917 & .944 \\
CamoDiff-E \cite{sun2025conditional} & TPAMI25 & F & .851 & .042 & .878 & .936 & .861 & .028 & .895 & .942 & .817 & .019 & .883 & .943 & -- & -- & -- & --\\
COMPrompter \cite{zhang2025comprompter} & SCIS25 & F & .858 & .044 & .882 & .942 & .876 & .030 & \bf .907 & \bf .955 & .821 & .023 & .889 & .949 & .857 & .026 & .906 & .955 \\
\midrule
UCOS-DA \cite{zhang2023unsupervised} & ICCVW23 & U & .606 & .127 & .701 & .784 & .656 & .085 & .755 & .819 & .513 & .086 & .689 & .740 & .591 & .095 & .715 & .802\\
UCOD-DPL \cite{yan2025ucod} & CVPR25 & U & .740 & \bf .077 & .793 & .862 & .818 & .043 & .850 & \bf .923 & .763 & \bf .031 & .834 & \bf .916 & .825 & \bf .031 & \bf .864 & \bf .931\\
EASE \cite{du2025shift} & CVPR25 & U & \bf .771 & .078 & \bf .807 & \bf .865 & \bf .833 & \bf .039 & \bf .866 & .915 & \bf .833 & .039 & \bf .866 & .915 & .827 & .037 & .864 & .916\\
SdalsNet \cite{shou2025sdalsnet} & AAAI25 & U & .664 & .117 & .696 & .801 & .697 & .084 & .738 & .826 & .572 & .084 & .738 & .826 & .666 & .080 & .724 & .837\\
\midrule
SAM2 \cite{tang2024evaluating, ravi2024sam} & ArXiv24 & ZS & .648 & .128 & .744 & .830 & .658 & .059 & .796 & .842 & .156 & .062 & .759 & .785 & .680 & .090 & .764 & .807 \\
GenSAM \cite{hu2024relax} & AAAI24 & ZS & .659 & .113 & .719 & .775 & .662 & .057 & .781 & .851 & .681 & .067 & .765 & .838 & .580 & .090 & .764 & .807 \\
MMCPF \cite{tang2024chain}  & MM24 & ZS  & .683 & .105 & .737 & .772 & .676 & .055 & .797 & .847 & .669 & .064 & .756 & .820 & .631 & .059 & .737 & .807 \\
ProMaC \cite{hu2024leveraging} & NIPS24 & ZS & .725 & .090 & .767 & .846 & .777 & .059 & .815 & .884 & .716 & .042 & .805 & .876 & \color{blue} .790 & \color{blue} .044 & \color{blue} .833 & \color{blue} .899 \\
CAMF \cite{lei2025towards} & TPAMI25  & ZS & .729 & .105 & .788 & .814 & .818 & .054 & .861 & .882 & .717 & .055 & .808 & .832 & .720 & .067 & .805 & 817\\
UpGen \cite{du2025upgen} & TIP25  & ZS & .718 & .091 & .779 & .833 & .788 & .054 & .838 & .887 & .708 & .051 & .838 & .887 & .747 & .047 & .817 & .875\\
IAPF \cite{yin2025simple} & ArXiv25  & ZS & \color{green}.768 & \color{blue} .081 &  \color{green} .803 & \color{green} .864 & \color{blue} .828 & \color{blue} .043 & \color{blue} .862 & \color{blue} .916 & \color{blue} .799 & \color{blue} .033 &  \color{blue} .856 & \color{blue} .917 & -- & -- & -- & -- \\
RDVP-MSD \cite{yin2025stepwise} & MM25  & ZS & \color{red} .785 & \color{blue} .081 & \color{blue} .796 & .848 & .795 & .049 & .842 & .891 & .775 & .038 & .825 & .877 & \color{green} .814 & \color{green} .040 & .832 & \color{green} .904\\
QWen+SAM2 \cite{bai2025qwen2,ravi2024sam} & -  & ZS & .741 & \color{green} .085 & .790 & \color{blue} .849 & \color{green} .846 & \color{green} .037 & \color{green} .875 & \color{green} .924 & \color{green} .827 & \color{green} .027 & \color{green} .873 & \color{green} .930 & .785 & .045 & \color{green} .838 & .884\\
DSS & -   & ZS & \color{blue}.766 & \color{red} .078 & \color{red} .808 & \color{red}.870 & \color{red}.870 & \color{red}.031 & \color{red}.891 & \color{red}.940 & \color{red}.849 & \color{red}.022 & \color{red}.887 & \color{red}.942 & \color{red}.848 & \color{red}.034 & \color{red}.883 & \color{red}.926 \\
\bottomrule
\end{tabular}
\label{tab:perf_comparison}
\end{table*}
\textbf{Datasets and Evaluations Metrics.} We conduct experiments on four widely used COD datasets, including CHAMELEON (76 images)~\cite{skurowski2018animal}, CAMO-Test (250 images)~\cite{le2019anabranch}, COD10K-Test (2,026 images) ~\cite{fan2020camouflaged}, and NC4K (4,121 images) ~\cite{lv2021simultaneously}. Following previous work, mean absolute error ($\mathcal{M}$) ~\cite{perazzi2012saliency}, structure measure ($S_\alpha$)~\cite{fan2017structure}, mean E-measure ($E_\phi$)~\cite{fan2018enhanced} and weighted F-measure ($F^w_\beta$)~\cite{margolin2014evaluate} are adopted to evaluate our method.\\
\noindent\textbf{MLLM-SAM Baseline.} For comparison, we include a simplified Vision–Language guided Object Segmentation (VLOS) baseline following the two-stage pipeline commonly used in prior works. Given an input image $I$, a  MLLM first localizes potential camouflaged objects under a task-generic prompt $P_{\mathrm{VLOS}}$:
\begin{equation}
\mathcal{B}_{vl} = \text{MLLM}(I, P_\mathrm{VLOS}), \quad \mathcal{B}_{vl} \in \mathbb{R}^{n\times4}.
\end{equation}
Each bbox serves as a prompt for SAM to produce a corresponding mask, and all masks are subsequently merged to form a unified segmentation:
\begin{equation}
  \mathrm{m} = \bigvee_{b_i \in \mathcal{B}_{vl}} \mathrm{VFM}(I, b_i),
\end{equation}
where $\bigvee$ denotes element-wise logical OR operation across predicted masks. This baseline reflects the standard MLLM-SAM pipeline without additional prompt refinement. Moreover, the mask generated by VLOS is included as one of the candidate masks in our final selection stage, demonstrating that our framework and VLOS are complementary and jointly improve segmentation performance. We compute the mean score of top-$K$ masks and define it as the score of mask from VLOS.\\
\noindent\textbf{Implement Details.} In our experiments, we choose QWen2.5-VL-Instruct~\cite{bai2025qwen2} as the MLLM and SAM2 (ViT-L) \cite{ravi2024sam} for segmentation. For the feature encoder, we deploy the DINOv2~\cite{oquab2023dinov2}. A dimensionality reduction to 16 dimensions using PCA is performed preceding the clustering step to improve computational efficiency. We adopts Leiden algorithm~\cite{traag2019louvain} over the widely used K-mean~\cite{hartigan1979algorithm} for clustering DINOv2 features, as it adapts to varying camouflage complexities by automatically determining the cluster number, avoiding the biases introduced by a fixed $K$ in K-means. For DSS, all experiments are conducted using the PyTorch framework, and we run inference on single NVIDIA GeForce RTX 3090 GPU with 24 GB of memory. We randomly sample 1,000 images from the training sets of COD10K and CAMO as a validation subset for hyperparameter tuning. All hyperparameters were empirically determined on this subset to balance performance and computation cost, and the same configuration was used across all experiments to ensure fairness and generalization. The hyperparameter settings are summarized in \cref{tab:hyperparameter}. Prompts are provided in the supplementary material~\cref{sec:prompt}.
\begin{figure*}[!t]
  \centering
  \begin{subfigure}{0.16\linewidth}
    \includegraphics[width=\textwidth]{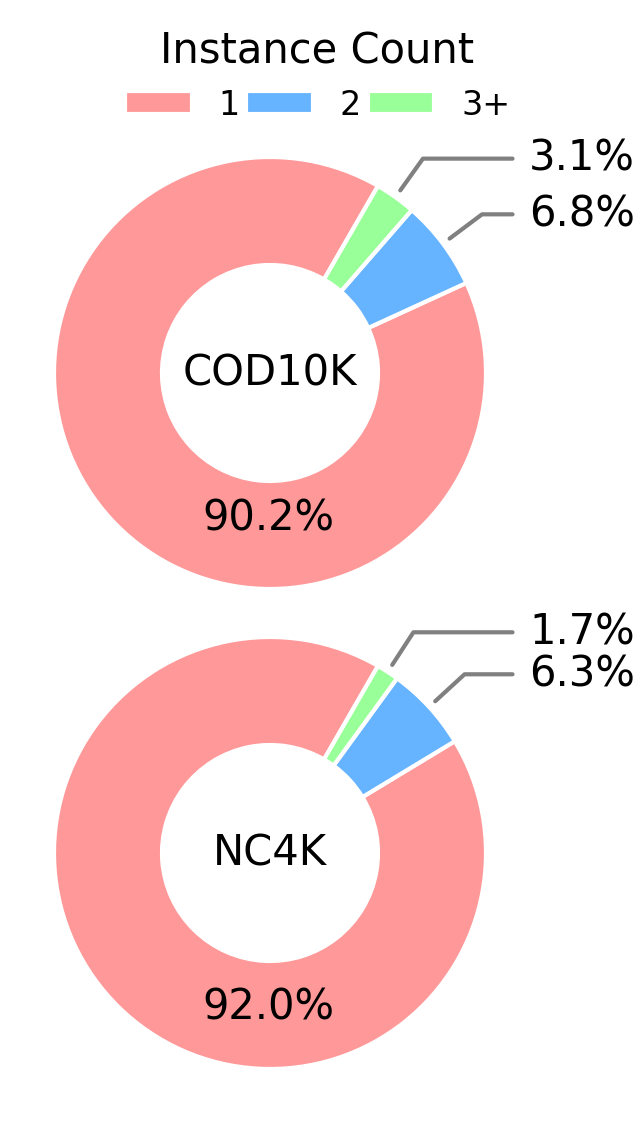}
    \caption{}
    \label{fig:pie}
  \end{subfigure}
  \begin{subfigure}{0.34\linewidth}
    \includegraphics[width=\textwidth]{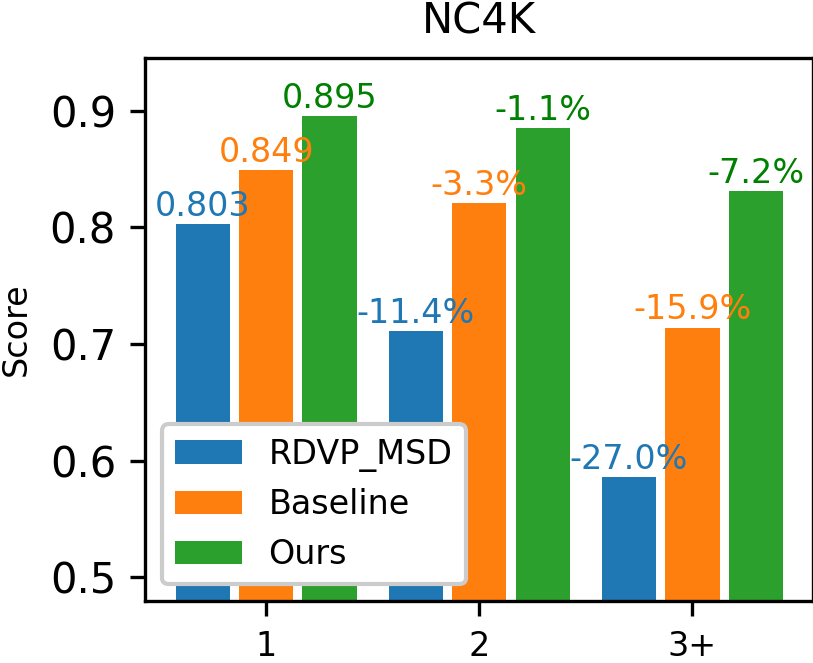}
    \caption{}
    \label{fig:cod10k_bar}
  \end{subfigure}
    \begin{subfigure}{0.46\linewidth}
    \includegraphics[width=\textwidth]{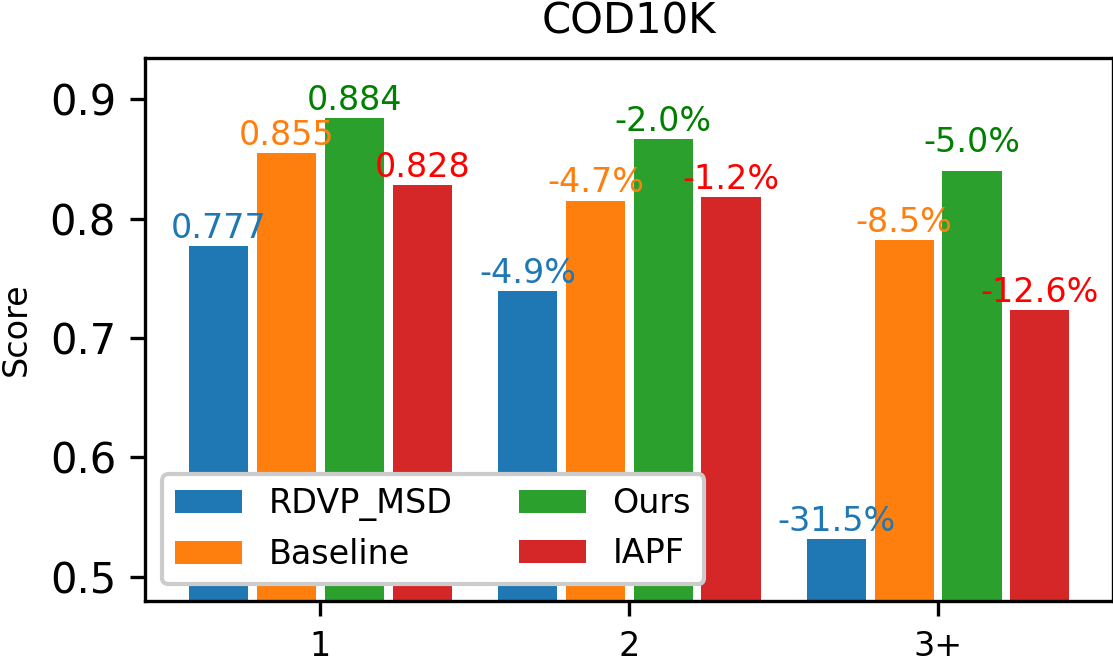}
    \caption{}
    \label{fig:nc4kk_bar}
  \end{subfigure}
  \caption{(a) Each pie chart shows the proportion of images containing different numbers of camouflaged objects. “3+” denotes images contain $(\geq3)$ camouflaged objects. (b) and (c) Performance comparison of different camouflaged instance counts on NC4K and COD10K. The y-axis is $score=\frac{S_\alpha+E_\phi+F_\beta^w}{3}-\mathcal{M}$, indicateing the overall performance.}
  \label{fig:bar}
\end{figure*}
\subsection{Comparison with State-of-the-Art Methods}
\textbf{Qualitative Analysis}. ~\cref{fig:pred_vis} qualitatively compares DSS with leading task-generic promptable segmentation methods. Our approach consistently produces more complete and compact segmentation. Specifically, DSS shows its strength in handling complex scenarios where MLLM-only localization solutions fail: (1) \textit{Under/Over-segmentation} (row 1-2). (2). \textit{Missed detection in multiple instances case} (row 3-4). (3) \textit{False detection} (row 5-6). In contrast, the proposed DSS is able to locate all concealed objects accurately, therefore producing high quality segmentations.\\
\noindent\textbf{Quantitative Comparision.} In~\cref{tab:perf_comparison}, we compare our proposed method with state-of-the-art fully-supervised, unsupervised, and zero-shot COD methods over four COD datasets. In Zero-shot paradigm, our DSS consistently outperforms all existing zero-shot methods across all datasets and evaluation metrics. The proposed method surpass all unsupervised methods without any form of training, underscoring its test-time adaptation capabilities. Furthermore, DSS narrows the performance gap with fully-supervised methods, highlighting its potential as a practical solution for COD tasks without the need for extensive annotated data.\\
\noindent\textbf{Performance Across Instance Counts.} We evaluate DSS's robustness in multi-instance (single, double, and three-or-more) scenarios despite their limited presence in real data, as shown by the instance distribution in \cref{fig:pie}. For each metric, we also list the relative change (\%) \textit{w.r.t.} the single-instance subset. As shown in ~\cref{fig:cod10k_bar,fig:nc4kk_bar}, existing methods exhibit significant performance drops as the number of camouflaged objects increases, highlighting their limitations in complex scenarios. In contrast, the proposed DSS achieves the minimal performance degradation with the increasing number of camouflaged objects, while maintaining the best performance on the single objet segmentation. This demonstrates the effectiveness of our feature-coherent object discovery approach in accurately locating multiple camouflaged objects within a single image.\\
\noindent\textbf{Computational Efficiency.} As shown in~\cref{tab:comp_efficiency}. The total processing time is dominated by the SMS module (30.59s, 72.9\%). Notably, the FOD module required only 7.74s (14.3\%) despite its iterative label refinement process. While the total inference time of DSS is 2.32 times of RDVP-MSD, it brings significant accuracy improvements as shown in~\cref{{tab:perf_comparison}}. In terms of GPU consumption, DSS requires only 17.90 GB GPU memory, significantly less than ProMaC and RDVP-MSD, by employing a 7B QWen2.5 instead of the 13B LLaVA used in these counterparts. Overall, DSS demonstrates a strong trade-off between segmentation accuracy and computational demands, making it a practical choice for real-world COS applications.
\begin{table}
  \caption{Comparison of computational efficiency, reported as average inference time per image (s). For our DSS framework, the time is decomposed into its three stages: FOD, mask generation, and SMS. Results are reported on the CHAMELEON dataset.}
  \centering
  \label{tab:comp_efficiency}
  \footnotesize
  \begin{tabular}{l|ccc}
    \toprule
    & ProMaC & RDVP-MSD & DSS \\
    \midrule
    Time (s) & 130.49 & \textbf{18.05} & 41.96 (7.74, 3.63, 30.59)\\
    GPU (GB) & 32.85 & 31.51  & \textbf{17.90} \\
    \bottomrule
  \end{tabular}
\end{table}
\begin{table}
  \caption{Ablation comparison of different inference settings. “Leiden” and “PC” denotes that the initial Leiden clustering map or the refined clustering map is used in the FOD module for extracting bboxes; “VLOS” denotes that the mask produced by QWen+SAM2 is included for mask selection.}
  \label{tab:comp_ablation}
  \centering
  \footnotesize
  \setlength{\tabcolsep}{4pt}
  \begin{tabular}{cccc|ccc}
    \toprule
    ID & Leiden & PC & VLOS & COD10K & CHAMELEON & CAMO \\
    \midrule
    1 & $\checkmark$ & $\times$ & $\times$ & .778 & .797 & .689\\
    2 & $\times$ & $\checkmark$ & $\times$ & .758 & .814 & .761\\
    3 & $\times$ & $\times$ & $\checkmark$ & .850 & .791 & .708 \\
    4 & $\checkmark$ & $\checkmark$ & $\times$ & .820 & .863 & .790\\
    5 & $\checkmark$ & $\times$ & $\checkmark$ & .886 & .872 & .801 \\
    6 & $\times$ & $\checkmark$ & $\checkmark$& .881 & .880 & .810\\
    7 & $\checkmark$ & $\checkmark$ & $\checkmark$ & \bf .892 & \bf .888 & \bf .823\\
    \bottomrule
  \end{tabular}
\end{table}
\subsection{Ablation Study}
To accurately evaluate the effectivenessess of PC and SBG module in isolation from the influence of the SMS, we adopt an Ideal Segmentation (Ideal Seg.) for comparison. Ideal Seg. refers to selecting, from all candidate masks, the one with the highest IoU score with the ground truth as the final mask. This allows us to evaluate the maximum achievable performance of our method when provided with perfect segmentation inputs. $score=\frac{S_\alpha+E_\phi+F_\beta^w}{3}-\mathcal{M}$ is used to indicate the overall performance.\\
\noindent\textbf{Effectiveness of PC.} Based on the ablation results in~\cref{tab:comp_ablation}, we draw several key observations. First, both the initial Leiden clustering (Row 1) and the refined clustering via the Part Composition module (Row 2) provide reasonable baselines, while their combination (Row 4) yields significantly better performance, demonstrating complementary advantages. Second, the discovery-segment baseline (Row 3) demonstrates inferior performance on the CHAMELEON and CAMO dataset, compared to the clustering-based methods (Row 4), highlighting the importance of visual feature guidance in COS. Finally, incorporating masks from VLOS and FOD (Row 7) achieves the best performance across all datasets, demonstrating that the proposed FOD module effectively complements the MLLM-SAM pipeline.
.\\
\noindent\textbf{Effectiveness of SBG.} To validate the effectiveness of our proposed SBG strategy, especitally on multiple-instance cases, we comparing performance based on bbox proposals generated from the self-similarity maps against those directly extracted from clustered and refined masks (Leiden+PC). As shown in~\cref{tab:SBG_ablation}, utilizing bboxes derived from similarity maps consistently outperforms the binary mask-based approach across both datasets. This improvement can be attributed to the enhanced completeness and accuracy of bboxes generated from similarity maps, as visually demonstrated in~\cref{fig:sim_vs_binary}. The results confirm that our SBG strategy effectively captures the nuanced features of camouflaged objects, leading to superior segmentation performance.\\
\noindent\textbf{Effectiveness of SMS.} To assess the effectiveness of the SMS module and the potential upper bound of our DSS framework, we adopt Ideal Seg. As shown in~\cref{tab:Ideal_seg}, the performance of our mask selection strategy surpass the QWen+SAM2 baseline, indicating that our mask selection strategy is effective in identifying the most relevant masks. However, the comparison with Ideal Seg. demonstates that there is still considerable room for improvement in mask selection strategy. This suggests that future work could focus on enhancing the reliability of mask evaluation and selection to further boost segmentation performance.
\begin{table}
  \caption{Ablation on SBG. The segmentation performance based on bboxes generated from: (1) Clustered and refined masks. (Leiden+PC). (2). similarity maps (Sim. Map). “$^{3+}$” denotes the subset of images containing three or more camouflaged objects.}
  \centering
  \setlength{\tabcolsep}{3.5pt}
  \label{tab:SBG_ablation}
  \footnotesize
  \begin{tabular}{l|cccc|cccc}
    \toprule
    \multirow{2}{*}{Methods} & \multicolumn{4}{c|}{NC4K$^{3+}$} & \multicolumn{4}{c}{COD10K$^{3+}$} \\
 & $F_\beta^{w}\uparrow$ & $\mathcal{M}\downarrow$ & $S_\alpha\uparrow$ & $E_\phi\uparrow$
 & $F_\beta^{w}\uparrow$ & $\mathcal{M}\downarrow$ & $S_\alpha\uparrow$ & $E_\phi\uparrow$ \\
    \midrule
    Leiden+PC & .795 & .058 & .812 & .897 & .723 & .034 & .794 & .881 \\
    Sim. Map& \textbf{.818} & \textbf{.054} & \textbf{.839} & \textbf{.917}  & \textbf{.745} & \textbf{.026} & \textbf{.821} & \textbf{.911} \\
    \bottomrule
  \end{tabular}
\end{table}
\begin{table}
  \caption{Ablation on the SMS module.}
  \centering
  \footnotesize
  \begin{tabular}{c|cccc}
    \toprule
    Method & CAMO & NC4K & COD10K & CHAMELEON\\
    \midrule
    VLOS & .708 & .845 & .850 & .791\\
    DSS & .737 & .869 & .871 & .852\\
    Ideal Seg. & \bf .823 & \bf .895 & \bf .892 & \bf .888\\
    \bottomrule
  \end{tabular}
  \label{tab:Ideal_seg}
\end{table}
\section{Conclusion}
In this paper, we introduced the Discover-Segment-Select (DSS) mechanism for zero-shot camouflaged object segmentation. The proposed framework begins with a visual-augmented discovery stage that generates high-quality region proposals through unsupervised feature clustering. These proposals are subsequently segmented into fine-grained masks and finally refined by a multimodal selection module to identify the optimal output. Extensive experiments demonstrate that our approach not only sets a new state-of-the-art in zero-shot camouflaged object segmentation, but also exhibits strong generalization capability—especially in challenging multiple-instance scenarios where it accurately localizes and segments numerous camouflaged objects within a single image. Moreover, the method maintains high computational efficiency, achieving superior accuracy with minimal GPU memory overhead. Ablation studies confirm the importance of each stage in the proposed pipeline. In the future, we plan to enhance the reliability of mask evaluation and incorporate multi-scale feature aggregation to better detect tiny camouflaged objects.

{
    \small
    \bibliographystyle{ieeenat_fullname}
    \bibliography{main}  
}

\clearpage
\setcounter{page}{1}
\maketitlesupplementary

\section{Boundary Contact Ratio}
\label{sec:impl_details_bc}
To quantitatively measure the extent to which a predicted mask touches the image boundary, we compute the boundary contact ratio $r_b\in[0,1]$. Given a binary segmentation mask $\mathrm{m}$ of size $H\times W$, we define an outer margin of width $n$ pixels along the four sides of the image. For each side, we count the number of foreground pixels that fall within this boundary strip, excluding overlapping corner regions to avoid double counting. Let $C_t, C_b, C_l, C_r,$ denote the number of non-zero pixels in these four edge regions, and $N_e$ be the total number of pixels within all edge strips. The boundary contact ratio is then defined as:
\begin{equation}
    r_b = \frac{C_t + C_b + C_l + C_r}{N_e}
\end{equation}
In our experiments, we set the margin width $n$ to 10 pixels. A higher $r_b$ indicates that the predicted object is more likely to be touching or extending beyond the image boundary, which is a common characteristic of background.
\section{Extended Ablation on SMS Module}
\label{sec:extend_ablation}
We conduct a quantitative study on MLLM selection accuracy, measuring how often different strategies choose the IoU-best mask. Also discuss efficiency and inference cost compared to simpler heuristics.

~\cref{tab:sms_ablation} compares several strategies for the final Semantic-driven Mask Selection (SMS) stage. For clarity, the methods are: (1) Heuristic Rules. Purely rule-based selection using scoring function defined in~\cref{eq:score}; (2) Single-pass Selection. Feed all candidate masks to the MLLM in a single prompt and let it select the best mask; (3) Top-$K$ Single-pass. Pre-filter candidates by heuristic scores and present only the top-$K$ masks to the MLLM in a single prompt; (4–6) Top-$K$ pairwise variants—iterative pairwise comparisons among the top-$K$ masks, differing in the order of comparisons (random, ascending/descending by heuristic score). Columns CAMO and CHAMELEON summarize the task performance under each selection strategy (higher is better). The Time column indicates the average inference time (s) per image for each strategy (lower is better).

From~\cref{tab:sms_ablation}, we observe distinct trends in both effectiveness and efficiency. The Heuristic Rules baseline achieves reasonable accuracy with minimal cost (1.14 s/image), demonstrating that spatial and consistency cues with boundary contactness offer a fast but approximate proxy for semantic quality. However, its rule-based nature limits adaptability in ambiguous cases where visual cues alone are insufficient. The Single-pass Selection strategy fails due to out-of-memory (OOM) errors, as feeding all masks at once results in extremely long multimodal prompts that exceed the context length of current MLLMs, highlighting a practical limitation. Although Top-$K$ Single-pass mitigates this, it performs poorly because presenting multiple masks in one query often causes semantic confusion for the MLLM. In contrast, pairwise strategies substantially outperform single-pass variants, confirming that decomposing the reasoning process into binary comparisons helps the MLLM make more consistent and discriminative judgments. Among them, the Top-$K$ pairwise (ascending) variant achieves the best accuracy across both datasets, as progressively eliminating low-quality masks guides the MLLM toward the most coherent result. However, this comes at the cost of increased inference time (~30 s/image) due to multiple MLLM calls. Overall, reasoning-based progressive selection, particularly with ascending order, provides the most robust balance between accuracy and efficiency, and is therefore adopted as our default SMS configuration.
\begin{table}[t]
    \caption{Comparison of different Mask Selection strategies. CAMO and CHAMELEON columnes measure selection accuracy (\%), while Time indicates average inference cost (seconds per image). “OOM” denotes out-of-memory errors.}
    \centering
    \footnotesize
    \setlength{\tabcolsep}{3pt}
    \begin{tabular}{c|ccc}
        \toprule
        Method & CAMO & CHAMELEON & Time (s)\\
        \midrule
        Heuristic Rules & 68.80 & 71.05 & \textbf{1.14}\\
        Single-pass Selection & OOM & OOM & OOM \\
        Top-$K$ Single-pass & 40.24 & 36.84 & 14.25\\
        Top-$K$ pairwise (random) & 54.00 & 59.21 & 31.14\\
        Top-$K$ pairwise (descending) & 50.80 & 43.42 & 30.81\\
        Top-$K$ pairwise (ascending) & \textbf{74.40} & \textbf{73.68} & 30.59 \\
        \bottomrule
    \end{tabular}
    \label{tab:sms_ablation}
\end{table}

\section{Prompt Design for MLLM}
\label{sec:prompt}
For reproducibility, we provide the exact prompt templates used in our experiments for both camouflaged object localization and mask selection tasks.\\
\textbf{Prompt for camouflaged Object localization.} Following \cite{tang2024chain}, physical and dynamic description of objects are included in the prompt to help MLLM better perceive the camouflaged objects. The prompt used to guide the MLLM in locating camouflaged objects and generating bounding boxes is shown in Figure~\ref{fig:prompt_cod}. \\
\noindent\textbf{Prompt for camouflaged Object localization.} The prompt used to instruct the MLLM to select the best mask from candidate masks is shown in~\cref{fig:prompt_sms}.

\begin{figure*}[t]
\centering
\begin{tcolorbox}[title={\normalsize Prompt Template for Camouflaged Object Localisation}, 
                  colback=promptbg, colframe=promptborder,
                  arc=2mm, boxrule=0.5pt]
\begin{lstlisting}[style=jsonstyle]
{
    "role": "user",
    "content": [
        {"type": "image", "image": img_path},
        {"type": "text", "text": """The image may contain a few animal/insect or human whose shape, color, texture, pattern and movement closely resemble its surroundings. Please identify them and provide their locations in the format of coordinates, as precisely as possible. The output should be in JSON format, e.g.:
        {
            "bbox_2d": [[x1, y1, x2, y2],[x1, y1, x2, y2]],
            "label": "dog"
        }
        DO NOT ADD ANY EXTRA INFO, JUST JSON!"""}
    ]
}
\end{lstlisting}
\end{tcolorbox}
\caption{Prompt template for camouflaged object localization using MLLM.}
\label{fig:prompt_cod}
\end{figure*}
\begin{figure*}[t]
\centering
\begin{tcolorbox}[title={\normalsize Prompt Template for Camouflaged Object Localisation}, 
                  colback=promptbg, colframe=promptborder,
                  arc=2mm, boxrule=0.5pt]
\begin{lstlisting}[style=jsonstyle]
{
    "role": "user",
    "content": [
        {"type": "text", "text": "The image is this."},
        {"type": "image", "image": f"data:image/png;base64,{img64}"},
        {"type": "text", "text": "Overlap from the original image through mask A is this."},
        {"type": "image", "image": f"data:image/png;base64,{best_maskb64}"},
        {"type": "text", "text": "Overlap from the original image through mask B is this."},
        {"type": "image", "image": f"data:image/png;base64,{maskb64}"},
        {"type": "text", "text": f"""
        CAMOUFLAGE MASK COMPARISON TASK
        IMAGE: The image may contain a few animal/insect or human whose shape, color, texture, pattern and movement closely resemble its surroundings.
        Overlap A: Current overlap mask
        Overlap B: New candidate overlap mask
        STEP-BY-STEP PROCESS:
        1. OBJECT IDENTIFICATION:
        - Carefully examine the image
        - Identify all hidden/concealed objects and their exact locations.
        2. SELECTION CRITERIA:
        - PRIMARY: Choose the mask that covers ALL identified objects completely
        - SECONDARY: Among masks that meet primary criterion, choose the one with least background
        - If no mask covers all objects, choose the one that covers the most objects
        OUTPUT JSON (DO NOT ADD ANY EXTRA INFO, JUST JSON!):
        {
            "better_mask": "Mask A" / "Mask B", 
        }
        """},
    ],
}
\end{lstlisting}
\end{tcolorbox}
\caption{Prompt template for mask selection using MLLM.}
\label{fig:prompt_sms}
\end{figure*}


\end{document}